\documentclass[]{cit_lab_mfr}

\usepackage{hyperref}
\usepackage{cleveref}
\usepackage{verbatim}

\usepackage{graphicx}
\usepackage{wrapfig}   
\usepackage{floatrow}
\usepackage{subcaption}
\usepackage{listings}
\usepackage{algorithm}
\usepackage[T1]{fontenc}

\usepackage{subcaption} %

\usepackage[toc,page,header]{appendix}


\definecolor{diy_pink}{RGB}{255,247,240}
\usepackage{multirow}          
\usepackage[table]{xcolor}     

\usepackage{graphicx}   
\usepackage{adjustbox}  
\usepackage{array}      
\usepackage{graphicx}
\usepackage{caption} 
\usepackage{xcolor}
\usepackage{pifont}
\newcommand{\xmark}{\ding{55}}  

\usepackage{booktabs}    
\usepackage{graphicx}    

\usepackage{subcaption}  

\usepackage{minitoc}

\title{\textbf{\textit{CamReasoner}}: Reinforcing Camera Movement Understanding via Structured Spatial Reasoning}

\author{%
\parbox{\textwidth}{\centering
Hang Wu$^{1}$ \qquad Yujun Cai$^{2, 3, \S}$ \qquad Zehao Li$^{4}$ \qquad Haonan Ge$^{1}$ \\[1mm]
Bowen Sun$^{1}$ \qquad Junsong Yuan$^{5}$ \qquad Yiwei Wang$^{1}$
}}

\affiliation{%
\parbox{\textwidth}{\centering\small
$^{1}$University of California, Merced \qquad $^{2}$University of Queensland \qquad $^{3}$Ant Group \\[1mm]
$^{4}$University of Chinese Academy of Sciences \qquad $^{5}$University at Buffalo, State University of New York \\[1mm]
}}

\contribution[\S]{Corresponding Author}

\renewcommand{\thefootnote}{\fnsymbol{footnote}}
\renewcommand{\thefootnote}{\arabic{footnote}}  

\abstract{
\begin{abstract}
/Understanding camera dynamics is a fundamental pillar of video spatial intelligence. However, existing multimodal models predominantly treat this task as a black-box classification, often confusing physically distinct motions by relying on superficial visual patterns rather than geometric cues. We present \textbf{CamReasoner}, a framework that reformulates camera movement understanding as a structured inference process to bridge the gap between perception and cinematic logic. Our approach centers on the Observation-Thinking-Answer (O-T-A) paradigm, which compels the model to articulate spatio-temporal observations and reason about motion patterns within an explicit reasoning block. To instill this capability, we construct a Large-scale Inference Trajectory Suite comprising 18k SFT reasoning chains and 38k RL feedback samples. To the best of our knowledge, \textbf{we are the first to employ RL for logical alignment in camera movement understanding}, ensuring motion inferences are grounded in structured visual reasoning rather than contextual guesswork. Built upon Qwen2.5-VL-7B, CamReasoner-7B improves binary classification accuracy from 73.8\% to 78.4\% and VQA accuracy from 60.9\% to 74.5\% over its backbone, consistently outperforming both proprietary and open-source baselines across multiple benchmarks.
\end{abstract}
}

\date{\today}
\checkdata[Corresponding]{\url{hangwu@ucmerced.edu}, \url{vanora.caiyj@gmail.com}}
\checkdata[Code]{\url{https://github.com/wuhang03/CamReasoner}}
\checkdata[Model]{\url{https://huggingface.co/LaurentWu/CamReasoner-7B}}

\begin{document}
\maketitle

\begingroup
\renewcommand{\thefootnote}{\fnsymbol{footnote}}
\endgroup

\section{Introduction}
Camera movement serves as the invisible brushstroke of cinematic storytelling, acting as a vital conduit to convey subtle emotions and orchestrate audience immersion. Whether through Scorsese's seamless tracking shots that create a seductive flow, or Spielberg's dynamic dollies that heighten suspense and awe, these trajectories represent the intentionality of a director shaping the viewer's perception of the physical and emotional world. In the landscape of Multimodal Large Language Models (MLLMs), camera movement understanding has emerged as a critical frontier that bridges the gap between passive pixel perception and active semantic reasoning. From the perspective of understanding, mastering these dynamics allows MLLMs to effectively decouple complex object movements from the observer's ego-motion, even when camera intrinsic parameters are unknown, thereby enabling a more robust comprehension of 3D spatial geometry and narrative intent. This transition from surface-level recognition to deep structural analysis moves an AI beyond surface-level pattern matching toward more structured motion analysis. Furthermore, from the perspective of generation, camera movement acts as the essential cornerstone for controllable video synthesis. As the frontier of AI shifts toward purposeful visual artistry, the ability to generate physically consistent videos relies fundamentally on the model's grasp of how trajectories interact with scene depth and perspective. By integrating these motion priors, MLLMs can move beyond disorganized pixel synthesis toward achieving true world modeling capabilities, where every shift in perspective is both physically grounded and narratively coherent.


Despite the growing demand for such sophisticated motion perception, current methodologies for camera movement understanding remain split between two constrained paradigms that struggle to meet these requirements. Traditional geometric approaches, such as SfM~\cite{schonberger2016structure, wang2024vggsfm} and SLAM~\cite{davison2007monoslam, Engel2014LSD}, rely on per-frame camera pose estimation to reconstruct trajectories, yet often require prior knowledge of camera intrinsics to maintain accuracy. These systems are fragile in dynamic environments because they often fail to disentangle the camera's ego-motion from the complex movements of subjects within the scene. In contrast, modern learning-based Video-Language Models (VLMs)~\cite{comanici2025gemini, qwen3vl, hurst2024gpt} offer a promising alternative by treating motion understanding as a semantic task. Leveraging vast pre-training corpora, these models achieve impressive alignment between visual cues and cinematic terminology, while even demonstrating an emerging capacity for high-level reasoning~\cite{liu2025shotbench}. However, these models predominantly operate as black boxes that leap from pixels to labels without explicit reasoning. This shortcut often triggers hallucinations because the model relies on contextual cues like a running athlete instead of decoding the actual physics of the lens. Consequently, they struggle to distinguish physically distinct yet visually similar operations such as a zoom versus a dolly. These limitations necessitate a framework that bridges the gap between raw perception and structured cinematic logic.

\begin{figure*}[t] 
  \centering
  \includegraphics[width=1.0\textwidth]{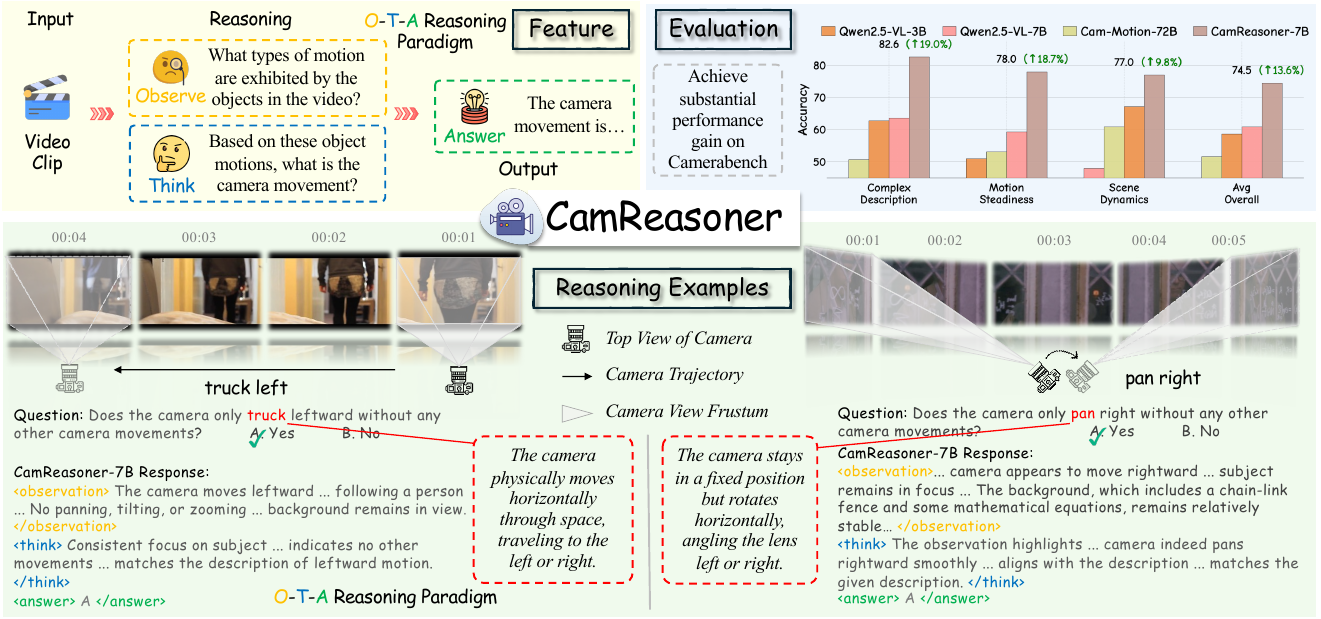}
  \caption{\textbf{Overview of CamReasoner}. We propose the \textit{<observation>}-\textit{<think>}-\textit{<answer>}paradigm to improve camera movement understanding. The figure illustrates our model generating detailed visual observations and logical thinking for movements like \textit{truck left} and \textit{pan right}.}
  \label{fig:teaser}
\end{figure*}

To bridge this gap, we introduce \textbf{CamReasoner}, a framework that reformulates camera movement understanding as a structured ego-motion inference task. Central to our approach is the Observation-Thinking-Answer (O-T-A) paradigm, which we implement through a sequential training pipeline of Supervised Fine-Tuning (SFT) and Reinforcement Learning (RL). To support this process, we constructed a Large-scale Inference Trajectory Suite comprising 18k SFT and 38k RL samples. While the SFT data transforms static labels into explainable reasoning chains to instill cinematic logic, the RL samples enforce logical alignment to ensure final verdicts are grounded in structured visual reasoning. By articulating detailed visual observations and reasoning about motion patterns, our model establishes a rigorous logical premise before delivering a verdict, substantially reducing reliance on contextual guesswork. To the best of our knowledge, we are the first to apply Reinforcement Learning for logical alignment in camera movement understanding. CamReasoner-7B achieves the best results among all compared methods, reaching 78.4\% in binary classification and 74.5\% in VQA. Notably, it excels in challenging Confusable Motion scenarios (60.7\%) through structured spatial reasoning. By integrating structural motion priors, CamReasoner provides a scalable path for MLLMs to achieve precise cinematic control and physically-consistent world modeling.

\noindent Our main contributions are summarized as follows:
\begin{itemize}
    \item \textbf{Structured Reasoning Paradigm:} We propose the Observation-Thinking-Answer (O-T-A) paradigm, reformulating camera movement understanding from black-box classification into a structured ego-motion inference task that decouples camera trajectories from dynamic scenes.

    \item \textbf{Inference Trajectory Suite:} We construct a Large-scale Inference Trajectory Suite with 18k SFT and 38k RL samples, transforming static labels into dense reasoning chains to instill spatio-temporal cinematic logic into MLLMs.

    \item \textbf{RL-driven Logical Alignment:} To the best of our knowledge, we are the \textbf{first to employ Reinforcement Learning} for logical alignment in camera movement understanding, achieving 78.4\% and 74.5\% accuracy in binary classification and visual question answering under our reformulated evaluation protocol, outperforming all compared baselines.
\end{itemize}

\section{Related Work}

\subsection{Camera Movement Understanding}

Camera movement understanding bridges geometric perception and semantic reasoning, yet current methods face significant bottlenecks. Traditional geometric approaches rely on SfM-based pseudo ground-truth, which is largely restricted to static scenes and ignores scene context~\cite{zhou2018stereo, ling2024dl3dv, jin2024stereo4d, schonberger2016structure}. Categorical recognition tasks, while providing semantic labels, often suffer from ambiguous taxonomy---confusing rotation with translation (e.g., pan vs. truck) and failing to capture co-occurring motions in complex cinematography~\cite{huang2020movienet, rao2020shot, argaw2022anatomy}. Recent studies further expose these weaknesses: Feng et al.~\cite{feng2026geometry} demonstrate that VideoLLM vision encoders weakly encode camera motion cues and propose injecting geometric signals from 3D foundation models to compensate, while VEU-Bench~\cite{li2025veu} reveals that Vid-LLMs struggle with editing-level understanding including shot motion, often performing below chance. Although recent benchmarks like ShotBench~\cite{liu2025shotbench, wu2025refineshot} evaluate nuanced cinematic grammar, existing methodologies lack a structured reasoning process to explain the visual cues driving their interpretations~\cite{hong2025motion, wang2024tarsier}. Consequently, there remains a critical need for a framework that integrates spatial evidence with logical deduction to achieve a holistic understanding across diverse video domains~\cite{zhou2018stereo, huang2020movienet, argaw2022anatomy, liu2025shotbench}.

\subsection{Multimodal Reinforcement Learning}

Inspired by the reasoning success of LLMs, recent works have extended structured thinking to MLLMs across diverse visual tasks~\cite{li2025star, sun2025reinforcement, feng2025video, sun2025spacevista, zhou2025reinforced, duan2025codeplot, chen2025advancing, chen2025ares, meng2025open}. These include task-specific reasoning models for image VQA~\cite{huang2025vision,mei2025surveycontextengineeringlarge,mei2025a1}, video understanding~\cite{feng2025video, li2025videochat}, object detection~\cite{yu2025perception, shen2025vlm}, segmentation~\cite{you2025seg}, and temporal grounding~\cite{wang2025time}. Notably, frameworks like SophiaVL-R1~\cite{fan2025sophiavl}, OneThinker~\cite{feng2025onethinker}, and FrameMind~\cite{ge2025framemind} introduce unified reasoning via reinforcement learning, while Thinking-with-Sound~\cite{xiong2025thinking} extends this paradigm to the audio domain through acoustic tool manipulation. However, despite the surge in RL-based multimodal reasoning, its application to camera movement understanding remains largely underexplored.

\section{Method}
We introduce a two-stage training strategy combining an SFT cold start with GRPO~\cite{shao2024deepseek} to instill the O-T-A reasoning paradigm into camera movement understanding. This pipeline ensures the model efficiently acquires structured reasoning traces while maintaining format adherence and task accuracy. To support this, we curate distinct datasets specifically tailored for sequential training.
\subsection{Data Collection and Curation}
High-quality, structured data is pivotal for teaching the model cinematic logic. We constructed two specialized datasets to facilitate the O-T-A paradigm across the SFT and RL stages.

\begin{figure*}[t] 
  \centering
  \includegraphics[width=1.0\textwidth]{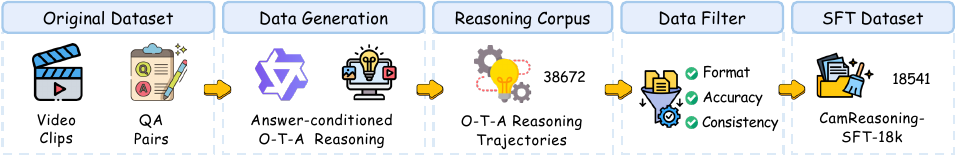}
\caption{\textbf{Data generation pipeline for CamReasoning-SFT-18k}. We utilize an answer-conditioned process to generate 38,672 initial trajectories, which are then filtered for format, accuracy, and consistency to retain 18,541 high-quality samples.}
  \label{fig:dataset}
\end{figure*}

\paragraph{CamReasoning-SFT-18k.} To instill structured reasoning, we curate the CamReasoning-SFT-18k dataset via an answer-conditioned synthesis pipeline. As shown in Fig.~\ref{fig:dataset}, we utilize Qwen2.5-VL-72B to generate 38,672 initial trajectories from original video-QA pairs, enforcing the \textit{<observation>}, \textit{<think>}, and \textit{<answer>} format. To ensure logical coherence, we implement a multi-dimensional filtration process using Qwen3~\cite{qwen3vl} to verify samples based on format adherence, label accuracy, and reasoning consistency. We further conduct human verification on a random subset of 500 samples, achieving an acceptance rate of 93\%, confirming the reliability of the automatic filtration. This results in 18,541 high-fidelity samples that serve as cold-start data for teaching the model step-by-step cinematic inference.

\paragraph{CamReasoning-RL-38k}
For the reinforcement learning stage, we construct the CamReasoning-RL-38k dataset to drive GRPO~\cite{shao2024deepseek}. Derived from the same CameraBench training split used for SFT data generation, this dataset comprises 38,451 video-QA pairs converted into our standardized instruction schema. Unlike the SFT phase, these samples lack reasoning annotations; instead, they serve as environmental queries for the model to refine its self-generated reasoning paths. This setup allows the model to optimize logical robustness and prediction accuracy through trial and error, guided by task-specific reward signals.

\subsection{Training Strategy}

Our training pipeline comprises two progressive stages designed to endow the model with robust O-T-A reasoning capabilities for camera movement understanding.
We first employ SFT as a cold-start initialization to adapt the model to the structured \textit{<observation>}-\textit{<think>}-\textit{<answer>}format.
Subsequently, we advance to a RL phase utilizing GRPO to further refine the reasoning logic and align outputs with task requirements.
Crucially, to mitigate the optimization instability inherent in single-task RL, we incorporate the EMA-GRPO strategy for robust advantage normalization.

\subsubsection{Supervised Fine-tuning}

We initiate the training process with SFT to bootstrap the model's capability in generating structured O-T-A (Observation-Think-Answer) reasoning traces.
As pre-trained MLLMs lack exposure to our specialized reasoning format encompassing \textit{<observation>}, \textit{<think>}, and \textit{<answer>} tags, SFT provides the essential cold-start initialization.
This phase instills adherence to the required formatting and logic patterns, ensuring the model can generate valid reasoning chains and camera movement predictions as defined in our dataset.

Formally, the SFT objective is to minimize the negative log-likelihood of the target sequence $o$ (which explicitly concatenates the observation, reasoning trace, and final answer). Given the input video $v$ and query $q$, the loss function is formulated as:
\begin{equation}
\mathcal{L}_{\text{SFT}}(\theta) = -\mathbb{E}_{(v, q, o) \sim \mathcal{D}_{\text{SFT}}} \left[ \sum_{t=1}^{|o|} \log \pi_\theta \left( o_t \mid v, q, o_{<t} \right) \right],
\end{equation}
where $o_t$ represents the $t$-th token of the target sequence $o$, and $o_{<t}$ denotes the preceding tokens.
$\pi_\theta$ is the policy of the multimodal model parameterized by $\theta$.
This stage enables the model to predict each subsequent token conditioned on the visual input, the query, and the generation history, establishing a robust foundation for the subsequent reinforcement learning stage.

\subsubsection{Reinforcement Learning Framework}

Building upon the SFT-initialized policy, we employ RL to further align the model's reasoning with task requirements. Given an input video clip $v$ and a textual query $q$, the policy $\pi_\theta$ generates a structured response $o$ comprising observation, reasoning trace, and predicted camera movement. The optimization objective is:
\begin{equation}
J(\theta) = \mathbb{E}_{o \sim \pi_\theta(\cdot|v, q)}[R(o)].
\end{equation}

\paragraph{Reward Design.}
We define a composite reward with two components: (1) a \textbf{Format Reward} $r_{fmt}$ that equals $1.0$ when the output strictly follows the \textit{<observation>}-\textit{<think>}-\textit{<answer>} structure and $0.0$ otherwise, and (2) an \textbf{Accuracy Reward} $r_{acc}$ that evaluates whether the prediction within \textit{<answer>} matches the ground truth, conditional on passing the format check. The final reward is:
\begin{equation}
\label{eq:reward}
R(o) = (1 - \lambda) \cdot r_{acc} + \lambda \cdot r_{fmt},
\end{equation}
where $\lambda$ balances structure adherence and prediction accuracy.

\paragraph{Optimization.}
We adopt GRPO to optimize the policy without the memory overhead of a value network. For each query, we sample a group of $G$ outputs from the current policy and estimate the baseline using the group average. The surrogate loss is:
\begin{equation}
\begin{aligned}
\mathcal{L}_{GRPO}(\theta) &= -\frac{1}{G} \sum_{i=1}^{G} \Bigg[ \min \Big( r_i(\theta) A_i, 
\text{clip}\big(r_i(\theta), 1\!-\!\epsilon, 1\!+\!\epsilon\big) A_i \Big) - \beta D_{KL} \Bigg],
\end{aligned}
\end{equation}
where $r_i(\theta) = \frac{\pi_\theta(o_i|q)}{\pi_{\theta_{old}}(o_i|q)}$ is the probability ratio, $\epsilon$ is the clipping parameter, and $\beta$ controls the KL penalty against the reference policy. To ensure training stability, we adopt EMA-GRPO~\cite{feng2025onethinker}, which replaces the batch-level standard deviation with an exponential moving average estimate $\sigma(t)$ for advantage normalization: $A_i(t) = ({R_i - \mathrm{mean}(\{R_j\})})/{\sigma(t)}$. This stabilizes gradient scaling and prevents spurious policy updates caused by low-variance rollouts.

\section{Experiments}

\subsection{Experimental Setup}

\paragraph{Benchmark and Dataset}
We evaluate on three benchmarks. \textbf{CameraBench}~\cite{lin2025towards} covers two tasks: 1) binary classification, where the model determines whether a specific motion primitive such as ``pan left'' is present in a given video, evaluated on 7,679 test samples; and 2) multi-choice VQA, where the model selects the correct camera movement description from several candidates, evaluated on 14,172 test samples spanning 9 categories. We modify the original benchmark by replacing mAP with accuracy for binary classification and reformulating VQA from pairwise confidence-score comparison into a standard multiple-choice format. These changes shift the evaluation focus from score calibration to decision-making correctness, enabling a more direct measure of whether a model can correctly interpret camera movements. For both tasks, we report the overall average accuracy weighted by the number of test samples in each category. Further details are provided in Appendix~\ref{sup_setting}. All baseline results are obtained by re-running each model under our reformulated protocol with identical prompts and decoding settings. \textbf{ShotBench}~\cite{liu2025shotbench} is a benchmark for cinematic shot understanding; we evaluate on its camera movement subset using accuracy. \textbf{RefineShot}~\cite{wu2025refineshot} extends ShotBench to assess reasoning reliability: an external LLM judges whether the model produces consistent reasoning across semantically equivalent queries, and the reliability score is the proportion of consistent outputs among all evaluated samples.

\paragraph{Implementation Details}
We evaluate all models under a unified protocol: video input is standardized to a maximum of 64 frames and 131,072 pixels ($128 \times 32 \times 32$) sampled at 1 FPS. The SFT phase uses LLaMA-Factory under DeepSpeed ZeRO-3, freezing the vision tower and multi-modal projector while fully fine-tuning the language model at a learning rate of $1 \times 10^{-5}$ with a cosine schedule for 3 epochs. The RL phase is implemented with verl at a learning rate of $2 \times 10^{-6}$ for 2 epochs, sampling $n{=}8$ rollouts per prompt with a low-variance KL penalty of $\beta{=}1 \times 10^{-2}$ and online filtering that prunes the top/bottom 1\% of reward samples. We set the $\lambda$ in Eq.~\ref{eq:reward} to 0.1 to balance different rewards. The policy is initialized from the second SFT epoch to balance reasoning capacity with exploration entropy. Full configurations and computational cost are in Appendix~\ref{sup:training}.

\subsection{Quantitative Results}

\definecolor{isabelline}{rgb}{0.96, 0.94, 0.90}
\definecolor{lightblue}{rgb}{0.85, 0.92, 1.0}
\begin{table*}[t!]
\centering
\setlength{\tabcolsep}{4pt}
\renewcommand{\arraystretch}{1.2}
\caption{\small \textbf{Binary classification performance} on camera movement understanding. Built upon Qwen2.5-VL-7B, \textbf{CamReasoner-7B} demonstrates significant improvements over its backbone and outperforms both proprietary and open-source baselines across most categories. Best results are in \textbf{bold} and second best are \underline{underlined}.}
\label{tab:binary}
\resizebox{\textwidth}{!}{
  \begin{tabular}{l @{\hspace{6pt}{\color{gray!60}\vrule width 0.3pt}\hspace{6pt}} cccccc @{\hspace{6pt}{\color{gray!60}\vrule width 0.3pt}\hspace{6pt}} cc @{\hspace{6pt}{\color{gray!60}\vrule width 0.3pt}\hspace{6pt}} cccccc @{\hspace{6pt}{\color{gray!60}\vrule width 0.3pt}\hspace{6pt}} c @{\hspace{6pt}{\color{gray!60}\vrule width 0.3pt}\hspace{6pt}} c}
    \toprule
    \multirow{2}{*}{\textbf{Model}} & \multicolumn{6}{c @{\hspace{6pt}{\color{gray!60}\vrule width 0.3pt}\hspace{6pt}}}{\textbf{Translation (Dolly/Pedestal/Truck)}} &
        \multicolumn{2}{c @{\hspace{6pt}{\color{gray!60}\vrule width 0.3pt}\hspace{6pt}}}{\textbf{Zooming}} &
        \multicolumn{6}{c @{\hspace{6pt}{\color{gray!60}\vrule width 0.3pt}\hspace{6pt}}}{\textbf{Rotation (Pan/Tilt/Roll)}} &
        \multirow{2}{*}{\textbf{Static}} &
        \multirow{2}{*}{\textbf{Avg}} \\
        \cmidrule(lr){2-7} \cmidrule(lr){8-9} \cmidrule(lr){10-15}
         & \textbf{In} & \textbf{Out} & \textbf{Up} & \textbf{Down} & \textbf{Right} & \textbf{Left} & \textbf{In} & \textbf{Out} & \textbf{Right} & \textbf{Left} & \textbf{Up} & \textbf{Down} & \textbf{CW} & \textbf{CCW} &  &  \\
        \midrule
    \rowcolor{isabelline}
    \multicolumn{17}{c}{\textit{Proprietary Models}} \\
    \midrule
    GPT-4o~\cite{openai2024gpt4ocard} & 63.6 & 61.6 & 67.2 & 53.5 & 69.6 & 59.4 & 76.0 & 57.1 & 64.5 & 76.2 & 63.1 & 77.0 & 60.0 & 74.5 & 74.7 & 66.5 \\
    \midrule
    \rowcolor{isabelline}
    \multicolumn{17}{c}{\textit{Open-source Offline MLLMs}} \\
    \midrule
    LLaVA-Video-7B~\cite{zhang2025llavavideo} & 35.5 & 46.1 & 31.4 & 53.5 & 48.9 & 41.2 & 48.4 & 22.6 & 15.2 & 21.9 & 20.3 & 23.7 & 44.7 & 59.6 & 32.9 & 37.7 \\
    InternVL2.5-8B~\cite{chen2025expandingperformance} & 44.3 & 55.2 & 61.9 & 66.0 & 46.4 & 66.0 & 61.8 & 65.3 & 63.9 & 55.3 & 68.0 & 64.5 & 63.6 & 50.9 & 52.0 & 59.0 \\
    Qwen2.5-VL-3B~\cite{bai2025qwen25vltechnicalreport} & 64.7 & 61.6 & \underline{84.7} & \underline{81.9} & 71.3 & \underline{79.9} & \textbf{85.1} & 72.3 & 77.9 & 79.5 & \underline{84.9} & \textbf{89.7} & 71.7 & 75.5 & 17.8 & 69.3 \\
    InternVL2.5-26B~\cite{chen2025expandingperformance} & 62.1 & 50.8 & 75.3 & 80.5 & 74.0 & 72.8 & 58.0 & 61.5 & 72.0 & 71.4 & 82.0 & 78.8 & 66.8 & 65.2 & 71.8 & 69.5 \\
    Qwen2.5-VL-7B~\cite{bai2025qwen25vltechnicalreport} & 62.7 & 56.8 & 82.8 & \textbf{82.8} & 75.8 & 79.7 & 58.7 & 68.6 & 72.9 & 80.1 & 83.8 & 86.0 & 67.9 & 73.8 & 74.2 & 73.8 \\
    Qwen2.5-VL-32B~\cite{bai2025qwen25vltechnicalreport} & \textbf{73.6} & \underline{76.1} & 80.2 & 80.2 & 72.5 & 65.2 & 67.6 & \underline{73.9} & 78.5 & 78.9 & 71.6 & 66.7 & \textbf{89.4} & \underline{78.3} & 78.0 & 75.4 \\
    \midrule
    \rowcolor{isabelline}
    \multicolumn{17}{c}{\textit{In-domain Trained Models}} \\
    \midrule
    Cam-Motion-7B~\cite{lin2025towards} & 39.7 & 48.6 & 32.6 & 64.2 & 50.5 & 54.5 & 51.3 & 14.2 & 17.6 & 18.0 & 12.0 & 20.8 & 47.2 & 75.1 & 35.0 & 38.8 \\
    Cam-Motion-72B~\cite{lin2025towards} & \underline{69.4} & \textbf{79.2} & 80.0 & 79.5 & \underline{77.6} & 77.5 & \underline{81.8} & \textbf{80.3} & \underline{80.5} & \underline{80.7} & 84.7 & \underline{87.5} & \underline{84.5} & \textbf{82.3} & \textbf{82.8} & \textbf{80.6} \\
    \midrule
    \rowcolor{lightblue}
    \textbf{CamReasoner-7B (Ours)} & 68.7 & 57.7 & \textbf{84.9} & 80.5 & \textbf{84.6} & \textbf{83.2} & 74.2 & 72.8 & \textbf{85.7} & \textbf{86.1} & \textbf{87.6} & 85.7 & 75.5 & 70.2 & \underline{80.1} & \underline{78.4} \\
    \bottomrule
  \end{tabular}
}
\end{table*}

\paragraph{Binary Classification.}
As presented in Tab.~\ref{tab:binary}, CamReasoner-7B achieves an overall accuracy of 78.4\%, surpassing GPT-4o by 11.9 percentage points and Qwen2.5-VL-32B by 3.0 points using a 7B backbone, and approaching Cam-Motion-72B within 2.2 points despite a 10× parameter difference. Compared to its backbone Qwen2.5-VL-7B, CamReasoner-7B yields a consistent average gain of 4.6 points, with the most pronounced improvements in Rotation and Zooming categories: \textit{Zoom In} increases from 58.7\% to 74.2\%, \textit{Pan Right} from 72.9\% to 85.7\%, and \textit{Move Right} from 75.8\% to 84.6\%. \textit{Static} recognition similarly improves from 74.2\% to 80.1\%, indicating enhanced capacity to classify stationary cameras in the presence of dynamic scene content. We further observe that the backbone exhibits a 4.0-point gap between \textit{Dolly In} and \textit{Zoom In}, suggesting limited ability to discriminate between the two visually similar approaching motions. Following O-T-A training, both categories improve to 68.7\% and 74.2\% respectively, with the gap reversed, which we attribute to the explicit spatial reasoning process encouraging finer-grained analysis of motion characteristics. A marginal regression is observed in \textit{Roll CCW}, which decreases from 73.8\% to 70.2\%, suggesting that the structured reasoning process may introduce slight uncertainty in rotation directions where the backbone already performs relatively well.

\definecolor{isabelline}{rgb}{0.96, 0.94, 0.90}
\definecolor{lightblue}{rgb}{0.85, 0.92, 1.0}
\begin{table*}[t]
\centering
\setlength{\tabcolsep}{4pt}
\renewcommand{\arraystretch}{1.2}
\caption{\small \textbf{VQA evaluation} on CameraBench across nine fine-grained dimensions of camera movement understanding. Built upon Qwen2.5-VL-7B, CamReasoner-7B consistently improves over its backbone across all categories and outperforms both proprietary and open-source baselines. Best results are in \textbf{bold} and second best are \underline{underlined}.}
\label{tab:vqa_evaluation}
\resizebox{\textwidth}{!}{
  \begin{tabular}{l @{\hspace{6pt}{\color{gray!60}\vrule width 0.3pt}\hspace{6pt}} ccccccccc @{\hspace{6pt}{\color{gray!60}\vrule width 0.3pt}\hspace{6pt}} c}
    \toprule
    \multirow{2}{*}{\textbf{Model}} & Motion \& & Scene & Motion & Motion & Confusable & Has & Shot & Only & Complex & Avg\\
    & Steadiness & Dynamics & Speed & Direction & Motion & Motion & Tracking & Motion & Description & Overall\\
    \midrule
    \rowcolor{isabelline}
    \multicolumn{11}{c}{\textit{Proprietary Models}} \\
    \midrule
    GPT-4o~\cite{openai2024gpt4ocard} & \underline{61.3} & 57.2 & 66.8 & 62.1 & 57.3 & \underline{68.4} & 58.5 & 47.6 & \underline{67.8} & \underline{64.5} \\
    \midrule
    \rowcolor{isabelline}
    \multicolumn{11}{c}{\textit{Open-source MLLMs}} \\
    \midrule
    Qwen2.5-VL-32B~\cite{bai2025qwen25vltechnicalreport} & 46.5 & 33.6 & 52.3 & 52.8 & 38.0 & 42.4 & 41.9 & 44.4 & 46.8 & 45.3 \\
    LLaVA-Video-7B~\cite{zhang2025llavavideo} & 42.8 & 44.3 & 46.8 & 48.4 & 26.7 & 54.9 & 52.4 & 34.0 & 59.3 & 52.3 \\
    InternVL2.5-8B~\cite{chen2025expandingperformance} & 44.1 & 55.2 & 49.1 & 51.5 & 27.3 & 56.0 & 53.9 & 36.3 & 60.6 & 54.1 \\
    Qwen2.5-VL-3B~\cite{bai2025qwen25vltechnicalreport} & 51.0 & \underline{67.2} & \textbf{76.4} & 56.4 & 40.7 & 58.3 & 58.1 & 45.3 & 62.8 & 58.6 \\
    InternVL2.5-26B~\cite{chen2025expandingperformance} & 56.4 & 39.4 & 54.3 & 62.9 & 36.0 & 56.2 & 61.1 & \underline{60.6} & 63.3 & 58.9 \\
    Qwen2.5-VL-7B~\cite{bai2025qwen25vltechnicalreport} & 59.3 & 48.0 & 65.2 & \underline{64.9} & 48.7 & 57.2 & \underline{63.5} & \textbf{64.2} & 63.6 & 60.9 \\
    \midrule
    \rowcolor{isabelline}
    \multicolumn{11}{c}{\textit{In-domain Trained Models}} \\
    \midrule
    Cam-Motion-7B~\cite{lin2025towards} & 45.3 & 54.0 & 46.3 & 23.2 & 21.0 & 46.9 & 50.7 & 36.2 & 37.6 & 40.5 \\
    Cam-Motion-72B~\cite{lin2025towards} & 53.1 & 60.9 & 48.3 & 52.4 & \textbf{62.7} & 50.1 & 49.5 & 58.1 & 50.7 & 51.6 \\
    \midrule
    \rowcolor{lightblue}
    \textbf{CamReasoner-7B (Ours)} & \textbf{78.0} & \textbf{77.0} & \underline{74.1} & \textbf{67.4} & \underline{60.7} & \textbf{71.8} & \textbf{71.3} & \underline{60.6} & \textbf{82.6} & \textbf{74.5} \\
    \bottomrule
  \end{tabular}
}
\end{table*}

\paragraph{VQA Evaluation.}
To further assess the model's ability to reason over competing motion descriptions, we evaluate on the VQA task under our reformulated multiple-choice protocol. As shown in Tab.~\ref{tab:vqa_evaluation}, CamReasoner-7B achieves 74.5\% overall, improving over its backbone Qwen2.5-VL-7B by 13.6 percentage points and exceeding GPT-4o by 10.0 points. The Cam-Motion models score between 40.5\% and 51.6\%, notably below their binary classification results. As discussed in Appendix~\ref{sup_setting}, the original CameraBench evaluates VQA via pairwise confidence-score comparison under a binary yes/no protocol, and the Cam-Motion models are optimized for this setting. Our reformulated protocol instead requires the model to select among competing motion descriptions, assessing explicit reasoning rather than relative score calibration. This shift inevitably disadvantages models trained exclusively under the original protocol, so we anchor our VQA analysis on comparisons with models that natively support multiple-choice output. Under this protocol, CamReasoner-7B demonstrates substantial improvements across most dimensions. The largest gains over the backbone appear on \textit{Scene Dynamics}, rising from 48.0\% to 77.0\%, \textit{Complex Description}, from 63.6\% to 82.6\%, and \textit{Motion \& Steadiness}, from 59.3\% to 78.0\%. The improvement on \textit{Complex Description} is particularly noteworthy, as the structured decomposition in the observation stage and synthesis in the thinking stage enable joint reasoning over multiple co-occurring motion attributes. The accuracy on \textit{Confusable Motion} at 60.7\% and \textit{Only Motion} at 60.6\%, while among the highest across all compared models, remains relatively modest, reflecting the inherent difficulty of distinguishing visually similar motion pairs under the multiple-choice setting.

\paragraph{Out-of-domain Generalization and Reasoning Reliability.}
To assess whether the acquired reasoning transfers beyond the training distribution, we evaluate on ShotBench~\cite{liu2025shotbench} and RefineShot~\cite{wu2025refineshot}, neither of which is included in our training data. As shown in Tab.~\ref{tab:bench_results}, CamReasoner-7B achieves 39.7\%, improving over Qwen2.5-VL-7B at 30.2\% by 9.5 percentage points. While the absolute accuracy reflects the challenging nature of ShotBench, the consistent improvement over the backbone suggests that the structured reasoning acquired through O-T-A training is not limited to the CameraBench distribution. We note that InternLM-XC2.5-7B at 38.8\% and InternVL3-14B at 38.2\% achieve comparable accuracy without in-domain training, indicating that further improving out-of-domain generalization remains an open challenge. Regarding reasoning reliability, CamReasoner-7B scores 96.1\%, slightly below the backbone's 98.9\%. This is expected, as the O-T-A paradigm produces substantially longer outputs, approximately 200 tokens versus fewer than 20 for direct-answer baselines, which increases the chance of minor inconsistencies across semantically equivalent queries. The reliability nevertheless remains comparable to InternVL3-14B at 97.1\%, confirming that the longer reasoning chains do not compromise overall logical coherence.

\begin{table}
\centering
\renewcommand{\arraystretch}{1.05}
\caption{\small \textbf{Performance and reliability} on \textbf{ShotBench} and \textbf{RefineShot}. CamReasoner achieves a +9.5\% accuracy gain over its Qwen2.5VL-7B backbone while maintaining high reasoning reliability (96.1\%).}
\label{tab:bench_results}
\scriptsize
  \begin{tabular}{l @{\hspace{8pt}{\color{gray!60}\vrule width 0.3pt}\hspace{8pt}} c @{\hspace{8pt}{\color{gray!60}\vrule width 0.3pt}\hspace{8pt}} cc}
    \toprule
    \textbf{Model} & \textbf{Size} & \textbf{Accuracy} & \textbf{Reliability Score} \\
    \midrule
    \rowcolor{isabelline}
    \multicolumn{4}{c}{\scriptsize\textit{Open-source MLLMs}} \\
    \midrule
    Iblip-vicuna-7B~\cite{dai2023instructblip}       & 7B  & 25.0 & - \\
    Iblip-vicuna-13B~\cite{dai2023instructblip}     & 13B & 22.0 & - \\
    VILA1.5-3B~\cite{lin2024vila}     & 3B  & 21.5 & - \\
    VILA1.5-8B~\cite{lin2024vila}    & 8B  & 36.9 & - \\
    VILA1.5-13B~\cite{lin2024vila}    & 13B & 31.3 & - \\
    LLaVA-NeXT-7B~\cite{li2024llava} & 7B  & 31.3 & - \\
    LLaVA-VID-7B~\cite{li2024llama}  & 7B  & 35.3 & - \\
    Ovis2-8B~\cite{lu2024ovis}       & 8B  & 35.3 & - \\
    InternLM-XC2.5-7B~\cite{dong2024internlm}   & 7B  & 38.8 & - \\
    InternVL2.5-8B~\cite{chen2025expandingperformance}    & 8B  & 33.6 & 96.3 \\
    InternVL3-2B~\cite{zhu2025internvl3e}      & 2B  & 38.1 & 95.4 \\
    InternVL3-14B~\cite{zhu2025internvl3e}     & 14B & 38.2 & 97.1 \\
    \midrule
    \rowcolor{isabelline}
    \multicolumn{4}{c}{\scriptsize\textit{Backbone \& Ours}} \\
    \midrule
    Qwen2.5VL-3B~\cite{bai2025qwen25vltechnicalreport}   & 3B  & 31.9 & 98.5 \\
    Qwen2.5VL-7B~\cite{bai2025qwen25vltechnicalreport}   & 7B  & 30.2 & 98.9 \\
    \rowcolor{lightblue}
    \textbf{CamReasoner-7B(Ours)} & 7B & 39.7 {\scriptsize\color{teal}(+9.5)} & 96.1 \\
    \bottomrule
  \end{tabular}
\end{table}

\begin{figure*}[t] 
  \centering
  \includegraphics[width=1.0\textwidth]{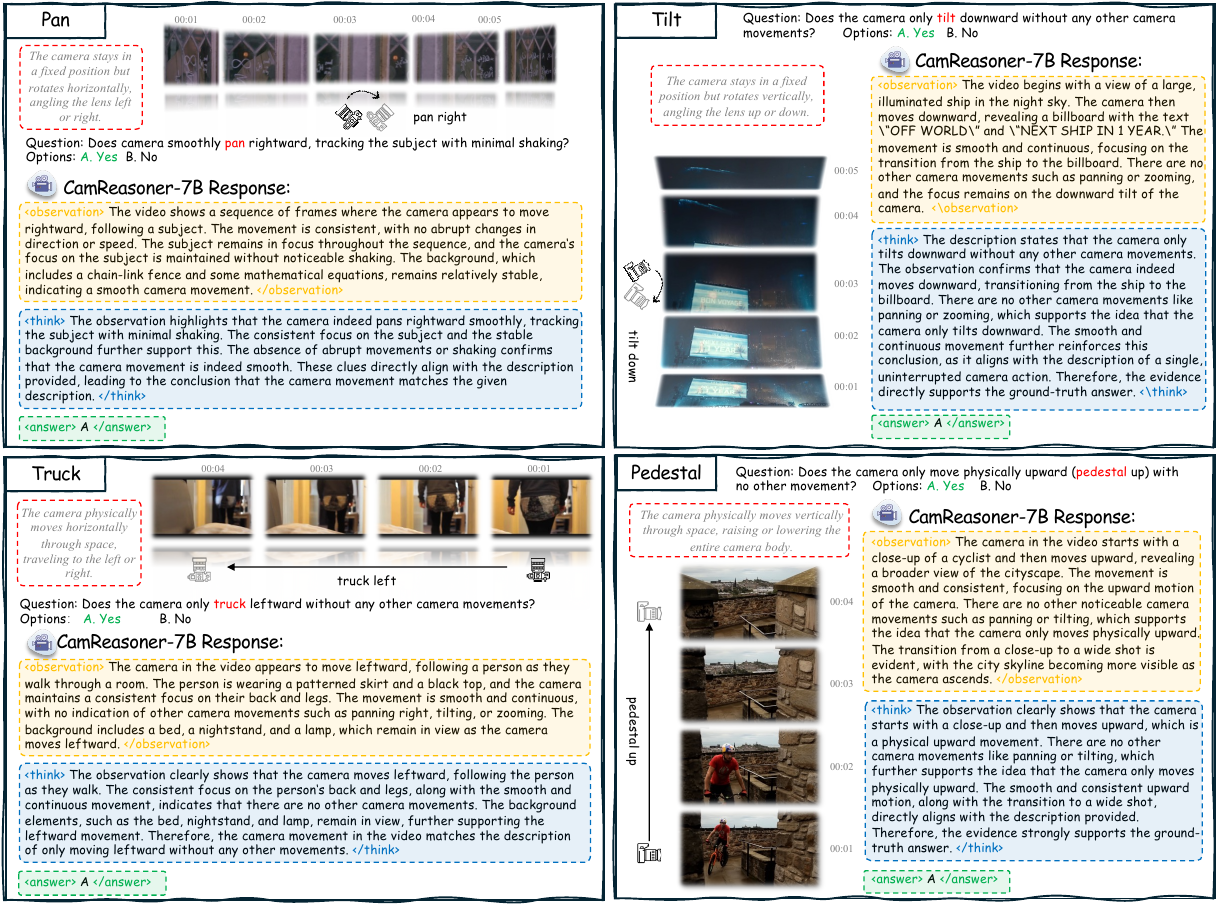}
  \caption{\textbf{Qualitative results} across four typical camera movements. For each case, we visualize the temporal frame sequence alongside the CamReasoner-7B response. The model demonstrates robust spatial reasoning by generating detailed \textit{<observation>} of visual cues and a logical \textit{<think>} process to accurately identify the movement and provide the final \textit{<answer>}.}
  \label{fig:results}
\end{figure*}

\subsection{Qualitative Results}
The qualitative results in Fig.~\ref{fig:results} substantiate the superior performance of CamReasoner-7B in camera movement reasoning. Through a structured \textit{<observation>}-\textit{<think>}-\textit{<answer>}pipeline, the model effectively translates fine-grained visual cues into logical deductions. 

During the observation phase, the model performs a detailed temporal scan to capture subtle shifts in subject-background relations. This granular perception allows it to decouple ego-motion from environmental noise, explicitly describing transitions between visual landmarks. In the subsequent thinking stage, CamReasoner engages in deductive verification grounded in cinematographic principles. By cross-referencing perceived displacements with the absence of confounding rotations, the model demonstrates that it has learned to produce structured reasoning chains that reference relevant spatial-temporal cues rather than relying on superficial statistical correlations. 

Finally, the alignment between these textual reasoning chains and the high accuracy underscores the efficacy of our training strategy. By refining a foundational reasoning checkpoint through reinforcement learning, we successfully bridge the gap between low-level visual perception and high-level semantic understanding.

\subsection{Training Curves}

\begin{figure}[t]
    \centering
    \includegraphics[width=1.0\linewidth]{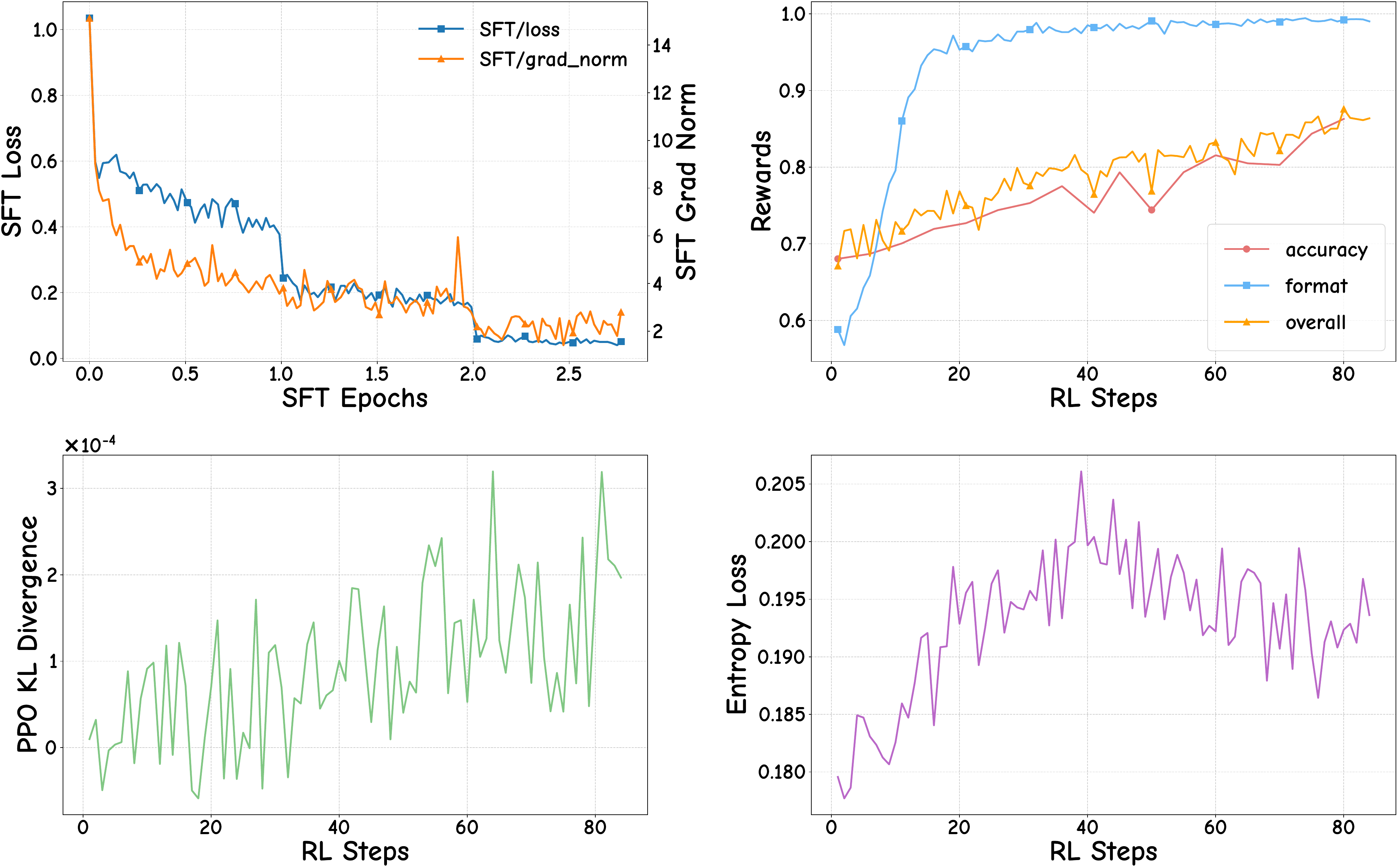}
    \caption{\textbf{Training curves for SFT and RL phases.} The top row shows the loss and grad\_norm during the SFT process; the bottom row visualizes the convergence of accuracy, format, and overall rewards during RL training.}
    \label{fig:rl_curve}
\end{figure}

The training dynamics of the SFT phase exhibit robust convergence and stability, as evidenced by the synchronized progression of loss and gradient norm in Figure~\ref{fig:rl_curve}. Specifically, the training loss undergoes a rapid initial decline and follows a distinct step-wise reduction at each epoch boundary, eventually stabilizing at approximately 0.05, which indicates that the model has effectively captured the underlying distribution of camera movement reasoning tasks. Throughout this process, the gradient norm remains within a well-behaved range between 2.0 and 4.0, ensuring a stable optimization landscape under the peak learning rate of $1.0 \times 10^{-5}$ and its subsequent cosine decay. Notably, we select the checkpoint from the conclusion of the second epoch as the initialization for the subsequent RL stage; at this juncture, the model has successfully acquired the fundamental paradigms of spatial reasoning and the \textit{<observation>}-\textit{<think>}-\textit{<answer>}output format, yet retains sufficient policy entropy by avoiding the potential overfitting characteristic of later training stages.

\begin{table*}[h]
\centering
\caption{\textbf{Ablation studies on CamReasoner.} We evaluate the contribution of each training stage and key RL hyperparameters.}
\label{tab:ablation_all}
\begin{subtable}[t]{0.48\textwidth}
    \centering
    \subcaption{\textbf{Training stages and RL initialization timing.} Bin.\ and VQA denote binary classification and visual question answering accuracy (\%), respectively.}
    \label{tab:ablation}
    \resizebox{\columnwidth}{!}{ 
    \begin{tabular}{lcccc}
    \toprule
    \textbf{Model} & \textbf{SFT} & \textbf{RL} & \textbf{Bin. Avg.} & \textbf{VQA Avg.} \\
    \midrule
    Base (Qwen2.5-VL-7B) & \xmark & \xmark & 73.8 & 60.9 \\
    \midrule
    \multicolumn{5}{l}{\textit{SFT at different epochs:}} \\
    \quad Epoch. 1 & \checkmark & \xmark & 74.5 & 65.8 \\
    \quad Epoch. 2 & \checkmark & \xmark & 75.3 & 69.5 \\
    \quad Epoch. 3 & \checkmark & \xmark & 75.8 & 69.2 \\
    \midrule
    \multicolumn{5}{l}{\textit{RL from different SFT checkpoints:}} \\
    \quad Epoch. 1 $\rightarrow$ RL & \checkmark & \checkmark & 76.2 & 71.2 \\
    \quad Epoch. 2 $\rightarrow$ RL (Ours) & \checkmark & \checkmark & \textbf{78.4} & \textbf{74.5} \\
    \quad Epoch. 3 $\rightarrow$ RL & \checkmark & \checkmark & 76.5 & 70.7 \\
    \bottomrule
    \end{tabular}
    }
\end{subtable}
\hfill
\begin{subtable}[t]{0.48\textwidth}
    \centering
    \subcaption{\textbf{RL hyperparameters:} reward weight $\lambda$, GRPO group size $G$, and advantage normalization (standard (Std.)\ vs.\ EMA-GRPO with momentum $\alpha$).}
    \label{tab:ablation_lambda_group}
    \vspace{0pt}
    \resizebox{\columnwidth}{!}{
    \renewcommand{\arraystretch}{1.1}
    \setlength{\tabcolsep}{10pt}
    \begin{tabular}{@{}lcccc@{}}
    \toprule
    \textbf{Reward weight $\lambda$} & 0.0 & \textbf{0.1} & 0.3 & 0.5 \\
    \midrule
    Bin. avg. & 74.6 & \textbf{78.4} & 75.9 & 71.4 \\
    VQA avg.  & 70.2 & \textbf{74.5} & 72.1 & 68.3 \\
    \midrule
    \textbf{GRPO group size $G$} & 2 & 4 & \textbf{8} & 16 \\
    \midrule
    Bin. avg. & 73.8 & 75.8 & \textbf{78.4} & 78.1 \\
    VQA avg.  & 70.6 & 72.3 & \textbf{74.5} & 74.2 \\
    \midrule
    \textbf{Adv. normalization} & Std. & $\alpha$=0.9 & $\boldsymbol{\alpha}$\textbf{=0.99} & $\alpha$=0.999 \\
    \midrule
    Bin. avg. & 75.1 & 76.8 & \textbf{78.4} & 77.6 \\
    VQA avg.  & 71.5 & 73.0 & \textbf{74.5} & 73.8 \\
    \bottomrule
    \end{tabular}
    }
\end{subtable}
\end{table*}

The RL phase, initialized from the second-epoch SFT checkpoint, further refines the model's policy to achieve rigorous logical alignment. As illustrated in Figure~\ref{fig:rl_curve}, the training dynamics demonstrate a consistent upward trend across all reward metrics. Specifically, the reward/format metric experiences a rapid surge within the first 20 steps, quickly stabilizing near a perfect score of 1.0. This indicates that the model swiftly masters the \textit{<observation>}-\textit{<think>}-\textit{<answer>}structural constraint, even under the exploration pressure of the RL environment. Concurrent with format stabilization, the \textit{reward/accuracy} exhibits a steady and robust climb from approximately 0.68 to over 0.85, proving that the GRPO-based optimization effectively sharpens the model's camera-centric reasoning boundaries. The synchronized growth of the \textit{reward/overall} metric, which eventually plateaus at a high equilibrium, confirms the successful balance between linguistic consistency and spatial-temporal accuracy. As shown in the qualitative results, this optimized policy enables CamReasoner-7B to execute meticulous deductive chains—such as explicitly verifying the absence of confounding motions in complex \textit{Truck} or \textit{Pedestal} shots—thereby bridging the gap between raw perception and logical cinematic analysis.

\subsection{Ablation Study}

We conduct ablation studies on CameraBench to evaluate the contribution of each training stage and key RL hyperparameters.
\paragraph{Training Stages and RL Initialization Timing.} As shown in Table~\ref{tab:ablation}, the base Qwen2.5-VL-7B achieves 73.8\% binary accuracy. SFT progressively improves performance, peaking at Epoch 3 with 75.8\%/69.2\%, though Epoch 3 shows signs of overfitting with a slight VQA decline compared to Epoch 2 (69.5\%). The choice of SFT checkpoint for RL initialization proves critical: Epoch 2$\rightarrow$RL achieves the best results (78.4\%/74.5\%), representing a 3.1\% binary improvement over SFT alone. Initializing from Epoch 3 degrades to 76.5\%/70.7\%, confirming that an overfitted policy lacks sufficient entropy for effective GRPO exploration, as the group-sampled outputs become nearly identical and collapse the advantage signal.
\paragraph{Reward Weight $\lambda$.} Table~\ref{tab:ablation_lambda_group} (top) shows that removing the format reward ($\lambda$=0.0) causes a notable drop to 74.6\%/70.2\% due to occasional malformed outputs. Performance peaks at $\lambda$=0.1, while larger values progressively degrade accuracy as the model over-prioritizes formatting at the expense of reasoning depth.
\paragraph{GRPO Group Size $G$.} Table~\ref{tab:ablation_lambda_group} (middle) shows that performance improves with larger groups, peaking at $G$=8 (78.4\%/74.5\%). Further increasing to $G$=16 yields no additional gains, suggesting eight samples already provide sufficiently stable advantage estimates.
\paragraph{Advantage Normalization.} Table~\ref{tab:ablation_lambda_group} (bottom) validates EMA-GRPO over standard batch-level normalization (75.1\%/71.5\%). The optimal momentum $\alpha$=0.99 achieves the best performance by filtering high-frequency noise while remaining responsive to the evolving reward distribution.

\section{Limitations}
Several limitations of CamReasoner stem from objective resource and data constraints.
First, due to computational budget, we validate CamReasoner only at the 7B scale; whether the O-T-A paradigm and RL-driven alignment yield comparable gains at larger (e.g., 72B) or smaller scales remains unexplored.
Second, our training data derives entirely from the CameraBench distribution, which has limited coverage of certain motion categories—particularly confusable pairs such as \textit{dolly} versus \textit{zoom}—restricting the model's exposure to diverse cinematic styles and domains; the 39.7\% absolute accuracy on the out-of-domain ShotBench reflects this data scarcity.
Third, the current pipeline lacks access to explicit geometric inputs such as monocular depth maps or camera intrinsics, which are not provided in existing video-QA benchmarks; integrating such signals could substantially improve disambiguation of visually degenerate motion pairs but requires either auxiliary annotation or additional foundation model inference at scale.

\section{Conclusions}
In this paper, we introduced \textbf{CamReasoner}, a framework that reformulates camera movement understanding as a structured ego-motion inference task. By employing the O-T-A paradigm, the model explicitly decouples camera motion from complex backgrounds through spatio-temporal reasoning. Our construction of a Large-scale Inference Trajectory Suite, comprising 18k SFT and 38k RL samples, provides a foundational resource for instilling cinematic logic into MLLMs. Experimental results demonstrate that CamReasoner-7B, built upon Qwen2.5-VL-7B, substantially improves over its backbone, raising binary classification accuracy from 73.8\% to 78.4\% and VQA accuracy from 60.9\% to 74.5\%, consistently surpassing both proprietary models such as GPT-4o and strong open-source baselines.  We hope CamReasoner can serve as a useful step toward bridging visual perception and structured deduction for camera movement understanding.

\clearpage

\bibliographystyle{plainnat}
\bibliography{main}

\clearpage
\appendix
\section{Details of Curated Dataset}

\begin{figure}[t]
    \centering
    \includegraphics[width=0.6\linewidth]{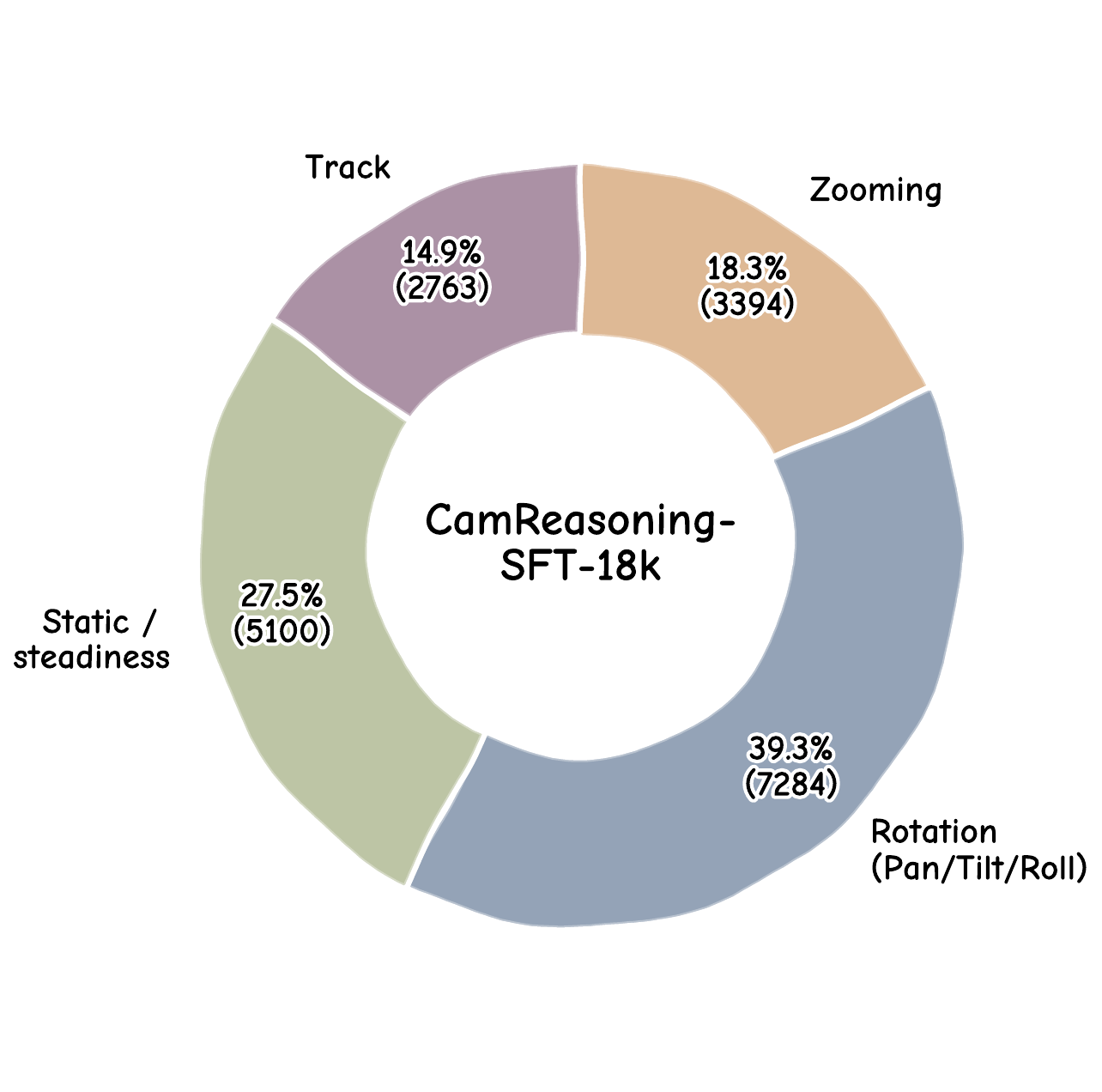}
    \caption{\textbf{Distribution of camera movement categories in CamReasoning-SFT-18k.} The dataset encompasses a diverse range of cinematographic motions, with a primary focus on dynamic rotations and stable static shots to foster robust reasoning capabilities.}
    \label{fig:dataset_stats}
\end{figure}

To equip \texttt{CamReasoner} with a comprehensive understanding of cinematography, we curated the \textbf{CamReasoning-SFT-18k} dataset, which consists of 18,541 high-quality samples. As illustrated in Fig.~\ref{fig:dataset_stats}, the dataset is categorized into four fundamental camera states to ensure balanced spatial-temporal learning:

\begin{itemize}
    \item \textbf{Rotation (39.3\%, 7284 samples):} The largest portion of the dataset covers rotational movements, including Pan, Tilt, and Roll. This allows the model to learn the nuances of angular perspective shifts.
    \item \textbf{Static / Steadiness (27.5\%, 5100 samples):} A significant subset is dedicated to fixed-camera scenarios. These samples are crucial for teaching the model to distinguish between subject movement and actual camera ego-motion.
    \item \textbf{Zooming (18.3\%, 3394 samples):} This category focuses on focal length changes (Zoom In/Out), aiding the model in perceiving depth-related visual cues.
    \item \textbf{Track (14.9\%, 2763 samples):} This includes translational movements such as Truck, Pedestal, and Dolly, completing the model’s mastery of the basic cinematographic degrees of freedom.
\end{itemize}

This curated distribution ensures that the model does not overfit to a single motion type. By providing a substantial number of \textit{Static} cases alongside \textit{Rotation} and \textit{Track} samples, we enforce the model's ability to perform "deductive verification"—explicitly confirming the absence of movement when visual cues are ambiguous.

\section{Details of Evaluation Setting}
\label{sup_setting}

\paragraph{Binary Classification} The original CameraBench evaluates binary classification using mean Average Precision (mAP) and ROC-AUC, which assess a model's ability to rank samples by calibrated confidence scores. However, under our reasoning-based paradigm, models produce explicit reasoning chains followed by a final Yes/No verdict, where the resulting token-level probabilities tend to be poorly calibrated for score-based ranking. In contrast, accuracy directly measures whether a model reaches the correct final decision through its reasoning process, which better aligns with our evaluation goal. We therefore adopt an accuracy-based evaluation protocol, where all models are evaluated under the same setting for fair comparison, with discriminative models thresholded at 0.5.

\paragraph{VQA Evaluation Reformulation.} The original CameraBench VQA evaluation pairs each question with a positive and a negative video, and determines correctness by comparing the model's yes/no confidence scores across the four possible (text, video) combinations. This protocol is designed to assess a model's discriminative ability by measuring whether it assigns relatively higher scores to correct pairings, which is effective for evaluating confidence calibration across paired samples. However, in our reasoning-based setting, models produce explicit reasoning chains that analyze visual evidence before arriving at a final answer, making pairwise score comparison less naturally applicable. We therefore reformulate the task as multiple-choice: given a single video, the model must select which description correctly matches the observed camera movement from a set of candidates. This formulation complements the original by shifting the focus from relative score discrimination to explicit reasoning over visual content, requiring the model to articulate why one description applies over another. Both formulations evaluate camera movement understanding from different perspectives; our reformulation is specifically designed to better capture the reasoning process that our O-T-A paradigm emphasizes. The reformulated task is evaluated with standard accuracy under the same protocol for all models.

\section{Training Details}
\label{sup:training}

\subsection{SFT Training}
Table~\ref{tab:hyperparams_sft} indicates the main hyperparameters used during SFT training phase.
\begin{table}[h]
\centering
\caption{Hyperparameter configuration for the SFT stage.}
\label{tab:hyperparams_sft}
\small
\begin{tabular}{ll}
\toprule
Hyperparameter & Value \\
\midrule
\multicolumn{2}{l}{\textit{Model}} \\
Base model & Qwen2.5-VL-7B-Instruct \\
Fine-tuning type & Full \\
Vision tower & Frozen \\
Multi-modal projector & Frozen \\
Language model & Trainable \\
\midrule
\multicolumn{2}{l}{\textit{Data}} \\
Max sequence length & 16,384 tokens \\
Max training samples & 20,000 \\
Image max pixels & 262,144 \\
Video max pixels & 16,384 \\
\midrule
\multicolumn{2}{l}{\textit{Optimization}} \\
Optimizer & AdamW \\
Learning rate & $1 \times 10^{-5}$ \\
LR scheduler & Cosine \\
Warmup ratio & 0.03 \\
Precision & bf16 \\
Training epochs & 3 \\
\midrule
\multicolumn{2}{l}{\textit{Infrastructure}} \\
Framework & LLaMA-Factory \\
Parallelism & DeepSpeed ZeRO-3 \\
Per-device batch size & 1 \\
Gradient accumulation steps & 2 \\
Checkpoint interval & 500 steps \\
\bottomrule
\end{tabular}
\end{table}

\subsection{GRPO Training}
Table~\ref{tab:hyperparams} indicates the main hyperparameters used during RL training phase.
\begin{table}[h]
\centering
\caption{Full hyperparameter configuration for RL training.}
\label{tab:hyperparams}
\small
\begin{tabular}{ll}
\toprule
Hyperparameter & Value \\
\midrule
\multicolumn{2}{l}{\textit{Data}} \\
Max prompt length & 16,384 tokens \\
Max response length & 4,096 tokens \\
Video FPS & 1.0 \\
Max pixels & 262,144 \\
\midrule
\multicolumn{2}{l}{\textit{Optimization}} \\
Optimizer & AdamW (bf16) \\
Learning rate & $2 \times 10^{-6}$ \\
LR warmup ratio & 0.0 \\
Weight decay & $1 \times 10^{-2}$ \\
Max gradient norm & 1.0 \\
\midrule
\multicolumn{2}{l}{\textit{RL Algorithm}} \\
Advantage estimator & EMA-GRPO \\
KL penalty type & low\_var\_kl \\
KL coefficient & $1 \times 10^{-2}$ \\
Rollouts per prompt ($n$) & 8 \\
Rollout temperature & 1.0 \\
Rollout top-$p$ & 1.0 \\
Online filtering & Top/bottom 1\% \\
Filter key & Accuracy \\
Reward weight $\lambda$ & 0.1 \\
\midrule
\multicolumn{2}{l}{\textit{Batch Sizes}} \\
Rollout batch size & 128 \\
Global batch size & 32 \\
Micro batch size (update) & 1 \\
Micro batch size (experience) & 1 \\
Validation batch size & 128 \\
\midrule
\multicolumn{2}{l}{\textit{Infrastructure}} \\
Precision & bf16 \\
FSDP & Full shard \\
Gradient checkpointing & Enabled \\
Padding free & Enabled \\
Dynamic batching & Enabled \\
Tensor parallelism & 4 \\
GPUs per node & 4 \\
Training epochs & 2 \\
\midrule
\multicolumn{2}{l}{\textit{Validation}} \\
Temperature & 0.7 \\
Top-$p$ & 0.95 \\
Samples per prompt & 1 \\
\bottomrule
\end{tabular}
\end{table}

\subsection{Computational Cost}
All training is conducted on a single node with 4$\times$NVIDIA H200 GPUs. The SFT stage takes approximately 3 hours for 3 epochs, and the GRPO stage takes approximately 7 hours for 2 epochs, totaling roughly 10 hours of wall-clock time. During inference, the O-T-A reasoning paradigm produces an average of approximately 200 tokens per sample, compared to fewer than 20 tokens for direct-answer baselines.

\end{document}